\documentclass[lettersize,journal]{IEEEtran}
\usepackage{amsmath,amsfonts}
\usepackage{multirow}
\usepackage{subfigure}
\usepackage[linesnumbered,ruled,vlined,onelanguage]{algorithm2e}
\usepackage{color}
\usepackage{array}
\usepackage[caption=false,font=normalsize,labelfont=sf,textfont=sf]{subfig}
\usepackage{textcomp}
\usepackage{stfloats}
\usepackage{url}
\usepackage{verbatim}
\usepackage{graphicx}
\usepackage{cite}
\begin{document}
	
	\title{ Spotlight Text Detector: Spotlight on Candidate Regions Like a Camera}
	\author{
		Xu~Han,  
		Junyu~Gao,~\IEEEmembership{Member,~IEEE,}
		Chuang~Yang, 
		Yuan~Yuan,~\IEEEmembership{~Senior Member,~IEEE}
		
		and Qi~Wang,~\IEEEmembership{~Senior Member,~IEEE}
		\thanks{
			
			X. Han, C. Yang are with the School of Computer Science, and with the School of Artificial Intelligence, Optics and Electronics (iOPEN), Northwestern Polytechnical University, Xi'an {\rm 710072}, China (E-mail: hxu04100@gmail.com, omtcyang@gmail.com).
			
			%
			%
			
		}
		\thanks{J. Gao, Y. Yuan, and Q. Wang are with the School of Artificial Intelligence, Optics and Electronics (iOPEN), Northwestern Polytechnical University, Xi'an {\rm 710072},  China (E-mail: gjy3035@gmail.com, y.yuan1.ieee@gmail.com, crabwq@gmail.com).}
\thanks{This work was supported by the National Natural Science Foundation of China under Grant U21B2041, 62306241. Qi Wang is the corresponding author.
}%

}


\markboth{{IEEE} Transactions on MULTIMEDIA}%
{Shell \MakeLowercase{\textit{et al.}}: A Sample Article Using IEEEtran.cls for IEEE Journals}


\maketitle

\begin{abstract}
		The irregular contour representation is one of the tough challenges in scene text detection. Although segmentation-based methods have achieved significant progress with the help of flexible pixel prediction, the overlap of geographically close texts hinders detecting them separately. To alleviate this problem, some shrink-based methods predict text kernels and expand them to restructure texts. However, the text kernel is an artificial object with incomplete semantic features that are prone to incorrect or missing detection. In addition, different from the general objects, the geometry features (aspect ratio, scale, and shape) of scene texts vary significantly, which makes it difficult to detect them accurately. To consider the above problems, we propose an effective spotlight text detector (STD), which consists of a spotlight calibration module (SCM) and a multivariate information extraction module (MIEM). The former concentrates efforts on the candidate kernel, like a camera focus on the target. It obtains candidate features through a mapping filter and calibrates them precisely to eliminate some false positive samples. The latter designs different shape schemes to explore multiple geometric features for scene texts. It helps extract various spatial relationships to improve the model's ability to recognize kernel regions.  Ablation studies prove the effectiveness of the designed SCM and MIEM. Extensive experiments verify that our STD is superior to existing state-of-the-art methods on various datasets, including  ICDAR2015, CTW1500, MSRA-TD500, and Total-Text.
\end{abstract}

\begin{IEEEkeywords}
	Text detection, arbitrary-shaped text, segmentation refinement.
\end{IEEEkeywords}
\indent
%
%
%
%
%
%
\section{Introduction}

\IEEEPARstart{S}{cene}  text detection aims to locate texts from images. It is a precondition for numerous computer vision tasks, including image retrieval, text recognition \cite{9695247}, \cite{9765383}, \cite{9798797}, autonomous driving, and text mining. As an essential step, scene text detection  \cite{leaftext}, \cite{rsmtd}, \cite{db}, \cite{dptext} has received increased focus in recent years. Different from general objects, scene text detection is more challenging as text enjoys varied scales, fonts, colors, and irregular shapes.

With the rapid development of object detection \cite{faster}, semantic segmentation, and instance segmentation, text detection has achieved tremendous progress recently. Existing scene text detection methods based on deep learning can be grouped broadly into three categories: regression-based \cite{textbox}, \cite{textbox++}, connected-component-based \cite{craft}, \cite{seglink}, and segmentation-based \cite{pan}, \cite{db++}. Compared to other methods, segmentation-based methods offer flexible pixel-level predictions, which are effective for handling text with arbitrary shapes. Unlike traditional semantic segmentation, text segmentation aims to obtain the contour of each instance individually with extra post-progressing after segmentation. The overlap of geographically close texts is the main challenging problem for these methods.

To cope with this problem, some methods \cite{pse}, \cite{pan}, \cite{db} predict text kernels and expand the prediction to restructure text instances, which are sensitive to the prediction of the text kernels. However, the text kernel is an artificial-defined concept that enjoys incomplete semantics. It is difficult for the model to predict accurately. On top of that, many patterns (such as fences, barks, and signs) in the background are also similar to the text texture, leading to missing detection and incorrect detection.
\begin{figure}[t]
	\centering
	\includegraphics[width=1.0\linewidth,scale=1.0]{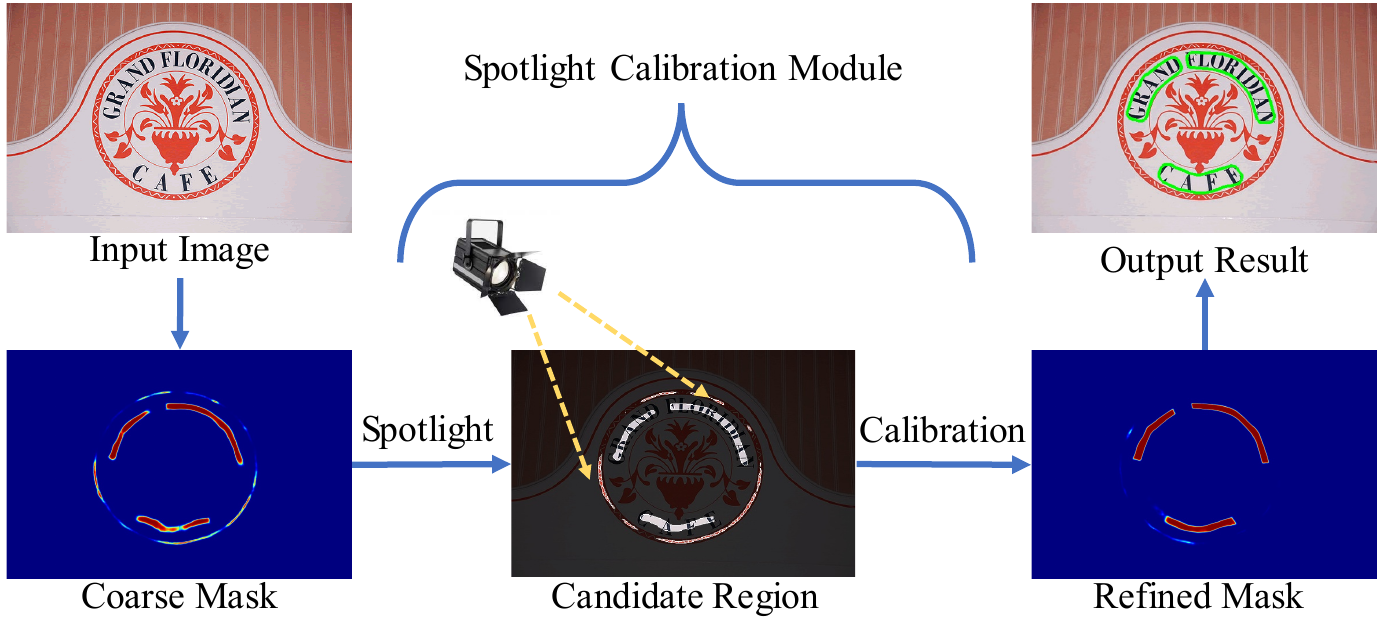}
	\caption{
		Illustration of spotlight calibration module. The coarse mask is generated like the existing method DB \cite{db}, PSE \cite{pse}, and PAN \cite{pan}. The module focuses on the feature of the candidate region and ignores others to calibrate further the prediction to generate the refined mask, which is superior to the coarse mask.  }\label{first}
\end{figure}
To address these problems, we propose a spotlight calibration module (SCM) that concentrates efforts on the candidate feature based on the coarse prediction, like a camera aimed at the target. 
Specifically, it eliminates some false positive samples by precisely calibrating the text kernel predictions and utilizes a cascade structure with a step-by-step increase receptive field to refine the extracting process of kernel features. The SCM effectively improves detection performance through the dual supervision of text kernels.
In addition, rechecking difficult-predicted objects aligns better with how humans cope with them. As shown in Fig. \ref{first}, existing methods only predict the coarse mask and many unstable predictions appear in the coarse mask due to the incomplete semantic features of text kernels. The SCM focuses on the feature of candidate regions based on the coarse mask and ignores the other regions to refine the judgment. Compared with the coarse mask, the refined mask is optimized further, which improves detection performance significantly.

Furthermore, different from general objects, the aspect ratio, scale, orientation, and shape of scene texts vary significantly, which needs to pay more attention to focus on them. Feature pyramid network (FPN) \cite{fpn}  is often found in various computer vision tasks to fuse different scale features. However, using FPN to segment various scale scene text effectively is challenging when adopting a lightweight backbone. Though PAN \cite{pan} proposes a feature pyramid enhancement module and a feature fusion module to fuse and enhance different scale features, its receptive field is limited and pays no attention to the various geometry features of scene texts. To address the above problem, we propose a multivariate information extraction module (MIEM) that explores multiple geometric features to cope with the diversity of scene texts in shape, scale, and orientation with a small amount of extra computation. Specifically, a parallel multiple geometry feature extraction structure is designed to force the model to capture various spatial relationships, which is lightweight and suitable for sophisticated scene texts.

Based on these modules, we propose an effective spotlight text detector (STD). It can detect arbitrary-shaped texts accurately, which reduces the influence of patterns like texts. The  contributions of this paper are summarized as follows:

\begin{enumerate}
	\item A spotlight calibration module (SCM) is proposed to 
	concentrate efforts on the candidate region based on the coarse mask, like a camera focus on the target. Moreover, it separates candidate features from input through a mapping filter and calibrates them to achieve dual supervision, which eliminates some false detection results, effectively improving detection performance.

	\item A multivariate information extraction module (MIEM) is proposed to explore multiple geometric features to address the diversity of texts in shape, scale, and orientation.   It extracts features on different receptive fields with a parallel multiple spatial structure feature extraction scheme to capture various spatial relationships.

	\item  An effective spotlight text detector (STD)  is proposed based on the above modules, which detects scene texts accurately while maintaining a competitive speed. Extensive experiments prove that our method achieves state-of-the-art (SOTA) performance on multiple public datasets, which include Total-Text, CTW1500, ICDAR2015, and MSRA-TD500.
\end{enumerate}

The rest structure of the paper is shown as follows. In Section \ref{related}, the related works about scene text detection are reviewed. Section \ref{method} describes the details of SCM and MIEM. Moreover, we analyze the impact of multiple combinations of loss. The ablation study on four public benchmarks proves the effectiveness of the  SCM and MIEM in Section \ref{experiment} further. Extensive experiments demonstrate the proposed STD is superior to existing methods. Finally, the conclusion of this paper is shown in Section \ref{conclusion}.
\section{Related work}
\label{related}
With the rapid development of deep learning, scene text detection obtains great progress. Existing methods can be divided into three classes: regression-based, connected-component-based, and segmentation-based.

\subsection{Regression-based methods}
Regression-based methods are generally inspired object detection methods with regression boxes, such as Faster-RCNN \cite{faster}. RRPN \cite{rrpn} adopted a modified pipeline based on Faster-RCNN, which uses a region-proposal-based approach to predict the orientation of the text instance. Liao \emph{et al.} proposed TextBoxes \cite{textbox}, which revises the shape of convolutional kernels and the default anchor of SSD \cite{ssd} to adopt the varied aspect ratio of scene texts. Based on it, TextBoxes++ \cite{textbox++} was proposed, which adds an angle parameter to detect multi-orientation scene texts. EAST \cite{east} classified the scene text as rotated and quadrangle to predict the bounding boxes or four corner points. RRD \cite{rrd} proposed a hierarchical inception module to extract multi-scale features and a text attention module to reduce the influence of background interference. Most of them need complex post-progress to recover the text instance, which limits their development.
Moreover, the above methods are unable to cope with the arbitrary-shaped text, which is common in scenes.  To address this problem, PCR \cite{pcr} proposed a progressive contour regression method to obtain contour evolution from horizontal to irregular shapes. FCE-Net \cite{fcenet} utilized the Fourier contour embedding to represent irregular-shaped text instances. ABC-Net \cite{abcnet} used a parametric representation based on the Bezier curve to represent arbitrary-shaped text instances. TextDCT \cite{textdct} utilized discrete cosine transform to represent text, which effectively approximates irregular-shaped texts. Although the above methods cope well with arbitrary-shaped texts, complex post-progressing limits efficiency.
\subsection{Connected-component-based methods}
These methods first detect the component or part of the text and then connect them to restructure the text.   CTPN \cite{ctpn} utilized a modified framework based on Faster-RCNN to detect the fixed-sized width text components and then connected them to reconstruct text instances. Textsnake \cite{textsnake} utilized a series of circles to represent text instances,  which are like the snake. To be specific, it predicted the center line, angle, and radius to model the circle. CRAFT \cite{craft} used the character-level annotations of synthetic images to extract the character features. Then, it predicted the affinity of characters to judge whether they belong to the same instance. SegLink \cite{seglink} predicted the segments and links of text instances that connect segments by the prediction of links. DRRG \cite{drrg} utilized the graph convolutional networks (GCN) to group the text parts of instances. Xu \emph{et al.} \cite{vrrca} designed a dense text segment representation method and a contour inference method to deal with arbitrary-shaped texts effectively.  MorphText \cite{morph} utilized the learnable deep module to replace the error-prone post-processing steps.  Although the above methods can represent arbitrary-shaped text well, complicated group progress influences accuracy and efficiency.

\begin{figure*}[t]
	\centering
	\includegraphics[width=1.0\linewidth,scale=0.7]{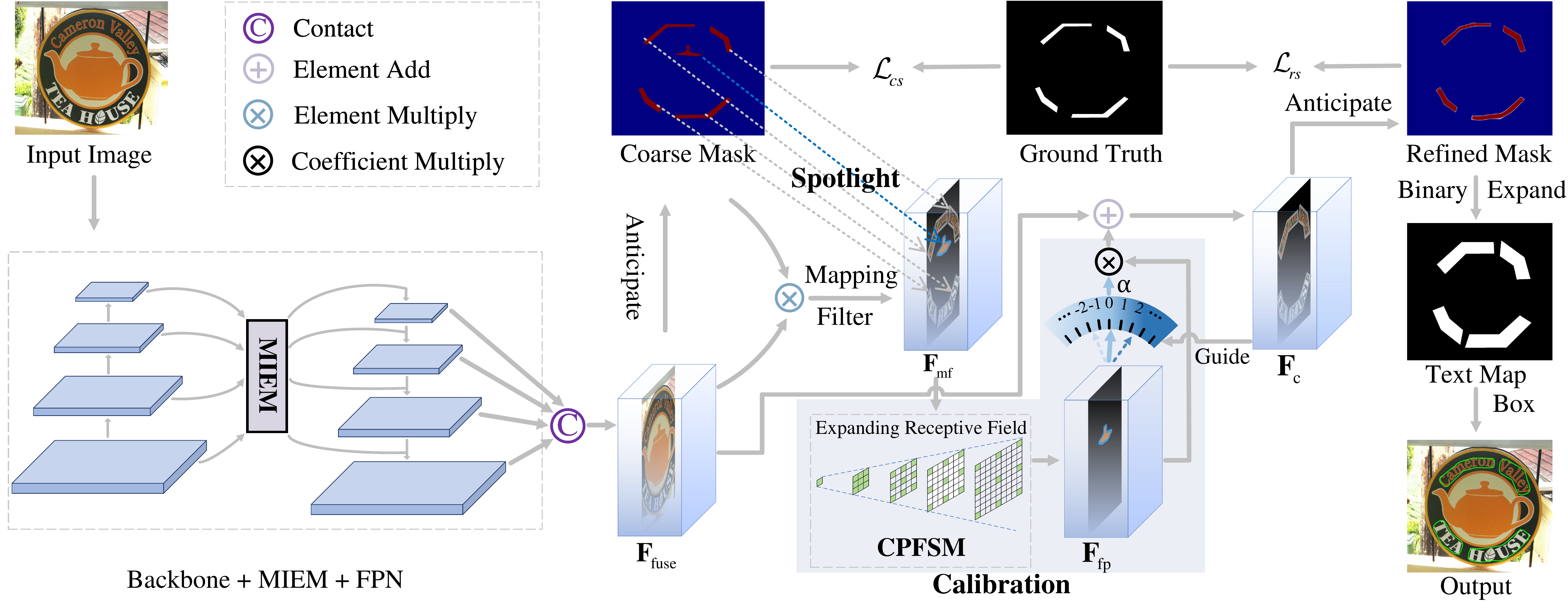}
	\caption{
		The overall structure of the proposed STD. It is composed of the backbone, multivariate information extraction module, feature pyramid network, coarse segmentation head, and spotlight calibration module. The MIEM and FPN are used to enhance the feature fusion. The SCM is used to calibrate the coarse mask to generate the refined mask. The cascading progressive feature search module (CPFSM) is a part of SCM, which is shown in Fig. \ref{pcm}. }\label{overview}
\end{figure*}

\subsection{Segmentation-based methods}

Segmentation-based methods adopt flexible pixel-level prediction and assist some extra predictions to recover arbitrary-shaped text instances. Pixellink \cite{pixellink} not only predicted the probability of pixels belonging to text regions but also predicted the relationship between pixels. Then, it grouped pixels to extract the instance contour by the prediction of pixel relationship. PSENet \cite{pse} segmented different scale kernels and utilized a progressive scaling method to expand them gradually to restructure text instances. TextField \cite{textfield} predicted a two-channel direction field and a segmentation map, which obtained final detection results via some morphological tools. DAST \cite{dast} improved the performance of detecting arbitrary-shaped text by estimating localization quality and using adaptive thresholds. As the complicated post-progressing, the above methods are unsatisfactory in speed. Subsequently, some real-time methods are proposed in succession. PAN \cite{pan} utilized a lightweight backbone to extract features and proposed a feature pyramid enhancement module and feature fusion module to compensate for the weak feature extraction. It predicted text kernels and similarity vectors to represent texts. DBNet\cite{db} adopted a simple post-processing to recover texts from kernels, improving efficiency significantly. Furthermore, it predicted the threshold to extract semantic features accurately. Based on it, DBNet++ \cite{db++} is proposed to recognize multi-scale semantic features, introducing a lightweight attention mechanism that slightly influences speed. Similarly based on DBNet \cite{db}, ADNet \cite{ad} and RSMTD \cite{rsmtd} focus on adaptively predicting the expansion distance in post-processing to improve the accuracy of the reconstructed results.  ZTD \cite{zoom} proposed two zoom strategy-based modules to alleviate feature defocusing and detail loss. RP-Text \cite{rp} proposed region context module and progressive module to extract text-related contextual feature and fuse multi-scale feature.  CM-Net \cite{cm} proposed a new text representation based on the concentric mask and a novel feature extraction module to extract text contours. TextBPN++ \cite{textbpn++} proposed a boundary transformer module and boundary energy loss to refine boundaries iterly and assist the learning of refinement. Most of the above methods reconstruct text instances based on text kernels that greatly cope with arbitrary-shaped text. However, the text kernel possesses incomplete high-level semantic features, and the presence of many patterns in nature with textures similar to text, which further exacerbates false positive predictions.

\section{Method}
\label{method}
The overall pipeline of the proposed STD is introduced first and illustrated in Fig. \ref{overview}.  Then, we describe the spotlight calibration module (SCM) and the multivariate information extraction module (MIEM) in detail. In addition, the loss function and label generation process are shown.

\begin{figure}[t]
	\centering
	\includegraphics[width=1.0\linewidth,scale=1.0]{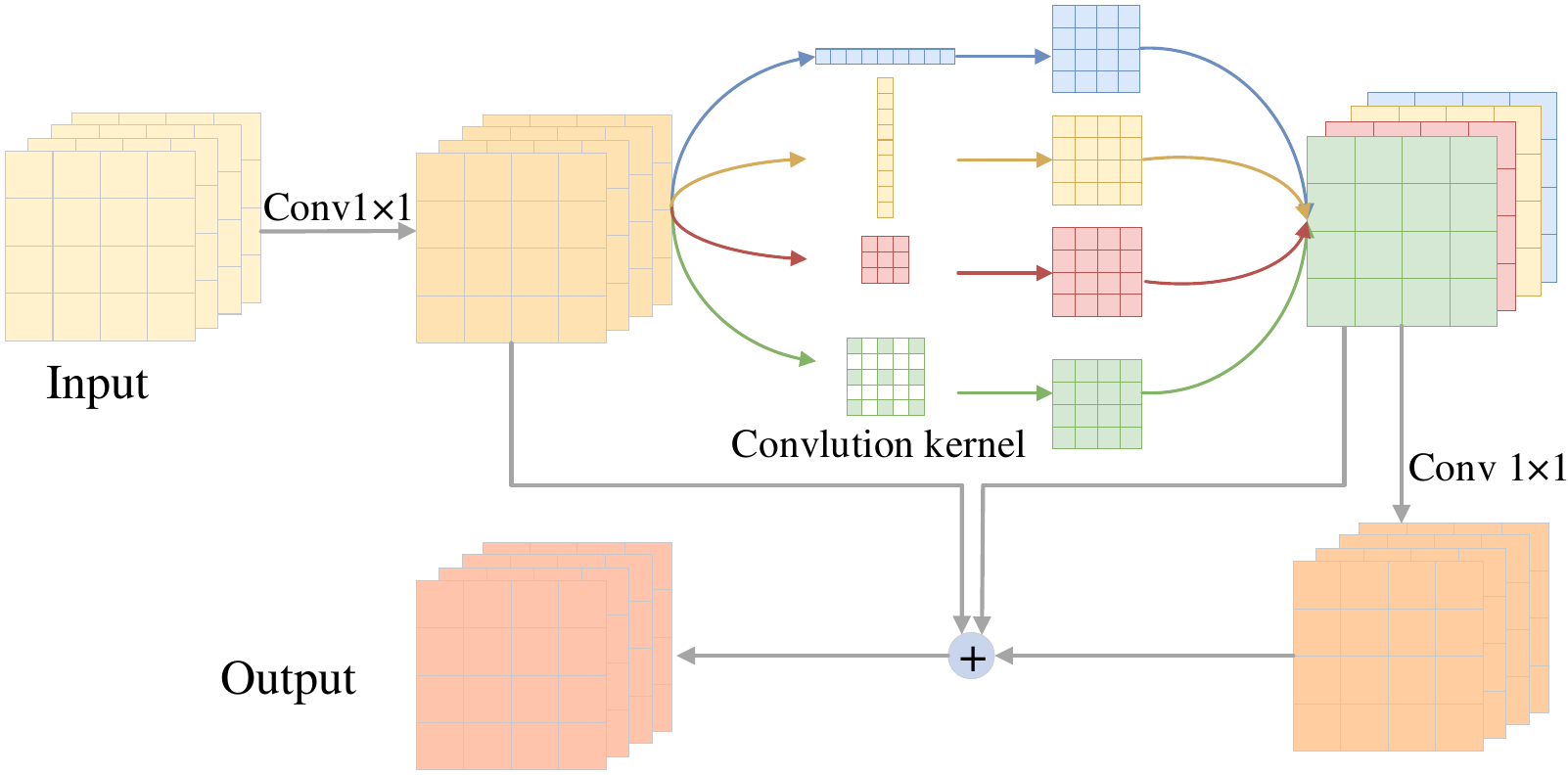}
	\caption{
		The overall structure of the proposed MIEM. It is used to extract multiple geometry features of text instances.}\label{miem}
	\label{miem}
\end{figure}
 \subsection{Overall Structure}
The overall structure of the proposed STD is shown in Fig. \ref{overview}. It includes the backbone, multivariate information extraction module (MIEM),  feature pyramid network (FPN) \cite{fpn}, coarse segmentation head, and spotlight calibration module (SCM). During the inference stage, different scale feature maps are extracted through the backbone. Then, MIEM is used to capture multiple geometry spatial features to cope with the diversity of texts in shape and orientation. The fused feature map $\mathbf{F_{fuse}}$ is gained through the feature pyramid network (FPN) \cite{fpn}, which includes fine-grained local and coarse high-level semantic features. Based on $\mathbf{F_{fuse}}$,  the coarse mask is predicted to select candidate regions and contrasts ground truth to optimize the predictions. The coarse mask segmentation head can be described as:
\begin{equation}
	\mathbf{W}^{\mathbf{1}}=\operatorname{conv}_{3\times 3}\left(\mathbf{F_{fuse}}\right),
\end{equation}
\begin{equation}
	\mathbf{M_{cs}}=\operatorname{Sigmoid} (\operatorname{conv}_{3\times 3}\left(\mathbf{W}^{\mathbf{1}}\right)),
\end{equation}
where $\mathbf{W}^{\mathbf{1}}$ represent hidden feature map. Subsequently, SCM refines the coarse mask, which checks the text kernel features to eliminate some false positive samples. The fused feature map $\mathbf{F_{fuse}}$ is multiplied by coarse predictions to obtain the mapping filtered feature map $\mathbf{F_{mf}}$, which spotlight the positive regions. Then, SCM calibrates $\mathbf{F_{mf}}$ to capture the false positive feature $\mathbf{F_{fp}}$. The calibration feature map $\mathbf{F_{c}}$  is obtained by adding the product of $\mathbf{F_{fp}}$ and $\alpha$ to  $\mathbf{F_{fuse}}$. Based on it,   a segmentation head is used to generate the refined prediction, which is binarized and expanded to obtain text regions. 
Finally,  the detection results are obtained through contour extraction.

\subsection{Multivariate Information Extraction Module}
Unlike general objects, scene texts exhibit significant variations in aspect ratio, scale, and shape. Segmentation-based methods, optimized for pixel-wise predictions, which require high-level semantic features more than other methods to cope with these problems. PAN \cite{pan} proposes the Feature Pyramid Enhancement Module (FPEM) and the Feature Fusion Module (FFM) to enhance text feature representation. These modules are both lightweight and effective. However, these modules overlook the diverse spatial features of scene texts. To address these challenges, we introduce the Multivariate Information Extraction Module (MIEM), which accommodates diverse variations in scene texts using distinct receptive fields. Additionally, MIEM extracts various geometric features to effectively handle the diversity in text shape and orientation, which introduces only a small amount of computation. Feature maps generated by the backbone have sizes  $\frac{H}{4}  \times \frac{W}{4} \times C$, $\frac{H}{8}  \times \frac{W}{8} \times 2C$, $\frac{H}{16}  \times \frac{W}{16} \times 4C$ and $\frac{H}{32}  \times \frac{W}{32} \times 8C$, where $C$, $W$ and $H$  are the channel number of the feature map and the width and height of the image. 
 \begin{figure}[t]
	\centering
	\includegraphics[width=1.0\linewidth,scale=1.0]{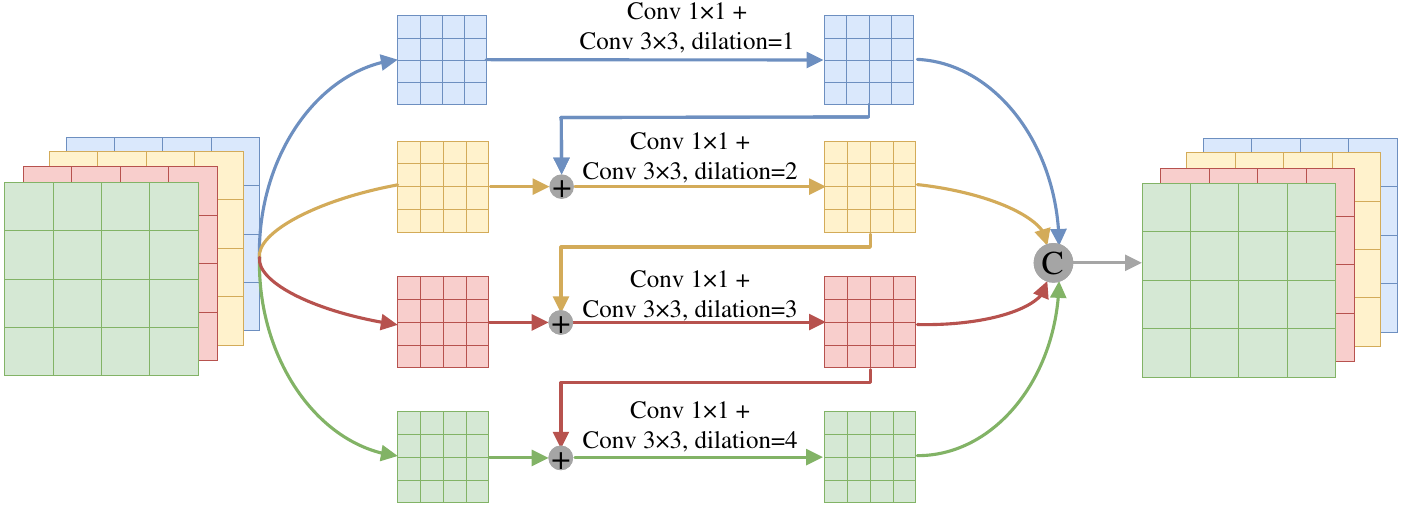}
	\caption{
		The overall pipeline of the Cascading Progressive Feature Search Module. It divides the feature maps into four groups and adopts a cascade scheme to obtain different receptive fields.}
	\label{pcm}
\end{figure}

As shown in Fig. \ref{miem}, for different scale feature maps, the process begins with a 1$\times$1 convolution layer to standardize the channels of the feature maps. Then, four groups of 1$\times$1 convolutions with Batch Normalization (BN) \cite{bn} and ReLU activation are performed on $\mathbf{{F}}_{\boldsymbol{b}}$
to reduce dimensionality and generate four distinct feature maps: ($\mathbf{{F}}_{\boldsymbol{b}}^1$, $\mathbf{{F}}_{\boldsymbol{b}}^2$, $\mathbf{{F}}_{\boldsymbol{b}}^3$, $\mathbf{{F}}_{\boldsymbol{b}}^4$). This step facilitates the extraction of various geometric features in the subsequent process. Subsequently, a multi-feature representation approach is employed to capture a variety of features from each of the corresponding feature maps. This process is further elucidated as follows: 

\begin{equation}
	\mathbf{\hat{F}}_{\boldsymbol{b}}^{\mathbf{1}}=\operatorname{conv}_{1\times 9,C^{'},\operatorname{dilation=1}}\left(\mathbf{F}_{\boldsymbol{b}}^1\right),
\end{equation}
\begin{equation}
	\mathbf{\hat{F}}_{\boldsymbol{b}}^{\mathbf{2}}=\operatorname{conv}_{9\times 1,C^{'},\operatorname{dilation=1}}\left(\mathbf{F}_{\boldsymbol{b}}^2\right),
\end{equation}
\begin{equation}
	\mathbf{\hat{F}}_{\boldsymbol{b}}^{\mathbf{3}}=\operatorname{conv}_{3\times 3,C^{'},\operatorname{dilation=1}}\left(\mathbf{F}_{\boldsymbol{b}}^3\right),
\end{equation}
\begin{equation}
	\mathbf{\hat{F}}_{\boldsymbol{b}}^{\mathbf{4}}=\operatorname{conv}_{3\times 3,C^{'},\operatorname{dilation=2}}\left(\mathbf{F}_{\boldsymbol{b}}^4\right),
\end{equation}
where $C^{'}$ is 1/4 of the channel number of $\mathbf{F}_{\boldsymbol{b}}$.

Its computation is equivalent to that of a general 3$\times$3 convolution layer, enabling the capture of multiple geometric features to enhance the robustness of the model. Moreover, it maintains approximately the same parameters as the combination of FPEM (Feature Pyramid Enhancement Module) and FFM (Feature Fusion Module), yet it outperforms the latter. Thanks to the parallel structure design, MIEM  is slightly faster than the combination of FPEM and FFM for the same amount of computation.
\begin{equation}
 	\mathbf{\hat{F}}_{\boldsymbol{b}} = {\rm Concat}(\mathbf{\hat{F}}_{\boldsymbol{b}}^{\mathbf{1}}, \mathbf{\hat{F}}_{\boldsymbol{b}}^{\mathbf{2}}, \mathbf{\hat{F}}_{\boldsymbol{b}}^{\mathbf{3}}, \mathbf{\hat{F}}_{\boldsymbol{b}}^{\mathbf{4}}),
 \end{equation}
where ``$ \rm Concat $'' represents the concatenate operator.
 Subsequently, another $1\times1$ convolutional layer is used to obtain $\mathbf{\bar{F}}_{\boldsymbol{b}}$ based on $\mathbf{\hat{F}}_{\boldsymbol{b}}$. The
 final feature map  $\mathbf{{\check{F}}}_{\boldsymbol{b}}$ is obtained by:
 \begin{equation}
 	\mathbf{\check{F}}_{\boldsymbol{b}} = \mathbf{F}_{\boldsymbol{b}}+ \mathbf{\bar{F}}_{\boldsymbol{b}}+\mathbf{\hat{F}}_{\boldsymbol{b}}.
 \end{equation}
 
Finally, the feature pyramid network (FPN) \cite{fpn} is utilized to obtain fused feature maps based on different scales $\mathbf{\check{F}}_{\boldsymbol{b}}$. The coarse mask is predicted based on this fused feature map.

\subsection{Spotlight calibration module}
Most segmentation-based methods rebuild text instances based on text kernels. However, the text kernel is an artificial geometric concept that lacks complete semantic features, and many text-like patterns are distributed in natural scenes. This significantly increases the challenge of scene text detection. Based on this, we propose a Spotlight Calibration Module (SCM), which calibrates the prediction of text kernels to eliminate some false-positive samples. SCM abandons non-candidate areas and focuses on the features of candidate areas.
Specifically, as shown in Fig. \ref{overview}, with the help of the coarse mask $\mathbf{M_{cs}}$, SCM applies mapping to filter the fused feature map $\mathbf{F_{fuse}}$, thereby obtaining the mapped filtered feature map $\mathbf{F_{mf}}$, which filters out negative features. The operation can be described as follows:
	\begin{equation}
	\mathbf{F_{mf}} = \mathcal{M}(\mathbf{F_{fuse}}, \mathbf{M_{cs}})
\end{equation}
where $\mathcal{M}$ represents the mapping function that applies the coarse mask $\mathbf{M_{cs}}$ to the fused feature map $\mathbf{F_{fuse}}$ to obtain the mapped filtered feature map $\mathbf{F_{mf}}$. This mechanism ensures that SCM effectively focuses on the features of candidate areas while eliminating irrelevant negative features, thereby improving the accuracy and robustness of scene text detection.

Then, the cascading progressive feature search module (CPFSM) is used to extract false positive feature map $\mathbf{F_{fp}}$ based on the $\mathbf{F_{mf}}$, which expands the receptive field  step by step to calibrate the predicted positive feature map. As described in Fig. \ref{pcm}, to reduce the amount of computation, we do not process the feature map directly, but rather divide the feature map $\mathbf{F_{mf}}$  into four groups ($\mathbf{F}_{\mathbf{mf}}^1$, $\mathbf{F}_{\mathbf{mf}}^2$, $\mathbf{F}_{\mathbf{mf}}^3$, $\mathbf{F}_{\mathbf{mf}}^4$) via the channel order.
Subsequently, a series of increased receptive field operators are utilized to enhance the feature. It can be described as follows:
\begin{equation}
	\mathbf{H}_{\boldsymbol{i}}^{\mathbf{1}}=\left\{\begin{array}{lc}
		\operatorname{conv}_{1 \times 1,64}\left(\mathbf{F}_{\mathbf{mf}}^i\right), & i=1 \\
			\operatorname{conv}_{1 \times 1,64}\left(\mathbf{F}_{\mathbf{mf}}^i\right)+\mathbf{H}_{\boldsymbol{i}-\mathbf{1}}^{\mathbf{2}}, & i=2,3,4
	\end{array}\right. ,
\end{equation}
\begin{equation}
	\mathbf{H}_{\boldsymbol{i}}^{\mathbf{2}}=\operatorname{conv}_{3 \times 3,64,\operatorname{dilation=i}}\left(\mathbf{H}_{\boldsymbol{i}}^{\mathbf{1}}\right), i=1,2,3,4,
\end{equation}

\begin{equation}
	\mathbf{F_{fp}}=\operatorname{conv}_{1 \times 1,64}(\text {Concat}\left(\mathbf{H}_{\boldsymbol{1}}^{\mathbf{2}}, \mathbf{H}_{\boldsymbol{2}}^{\mathbf{2}}, \mathbf{H}_{\boldsymbol{3}}^{\mathbf{2}}, \mathbf{H}_{\boldsymbol{4}}^{\mathbf{2}}\right)),
\end{equation}
where $\mathbf{H}^{\mathbf{1}}$ , $\mathbf{H}^{\mathbf{2}}$ represent the hidden feature maps. 
$\alpha$ is a trainable parameter that constrains the feature map $\mathbf{F_{fp}}$. It evolves with the network during training, starting with an initial value of -1.
To be specific,   $\mathbf{F_{fp}}$ multiply  $\alpha$ and adds $\mathbf{F_{fuse}}$ to generate calibration feature map $\mathbf{F_{c}}$, which can be described as follows:
 \begin{equation}
 	\mathbf{F_{c}}=\mathbf{F_{fuse}}+ \alpha \times \mathbf{F_{fp}}. 
 \end{equation}
Based on $\mathbf{F_{c}}$, a segmentation head is used to predict refined masks $\mathbf{M_{rs}}$, which can be described as:
	\begin{equation}
		\mathbf{W}^{\mathbf{2}}=\operatorname{convt}_{2\times 2}(\operatorname{conv}_{3\times 3}\left(\mathbf{F_{c}}\right)),
	\end{equation}
	\begin{equation}
		\mathbf{M_{rs}}=\operatorname{Sigmoid} (\operatorname{convt}_{2\times 2}\left(\mathbf{W}^{\mathbf{2}}\right)),
	\end{equation}
where $\mathbf{W}^{\mathbf{2}}$ and $\operatorname{convt}$ represent hidden feature map and transposed convolution operation.  Finally, a simple post-processing is utilized to get detection results.


\subsection{Label Generation}
In this paper, the text kernel label is utilized to optimize the proposed method. Each instance is represented by $n$ sample points, which are distinct in different datasets.
First, the coordinates of points are converted to a binary text map. Then, the Vatti clipping algorithm  \cite{vatti1992generic} is utilized to shrink texts to text kernels.
According to the perimeter and  area of the instance, the amount of shrinkage of the text is calculated as follows: 
\begin{equation}\label{2}
	\mathcal{S}_{i} = \frac{{A}_i\times(1-\gamma^2)}{{P}_i},
\end{equation}
where $\gamma$ represents the shrinkage factor which is set to 0.4. $\mathcal{S}_{i}$ is the shrinkage that the $i$th instance need to shrink. The area and the perimeter of the $i$th text are represented by $A_i$ and  $P_i$, respectively.

\subsection{Optimization Function}
In this paper, the proposed method utilizes two loss functions to optimize the model that includes the coarse mask loss $\mathcal{L}_{cm}$ and the refined mask loss  $\mathcal{L}_{rm}$. The full loss function is formulated as follows:

\begin{equation}
	\mathcal{L}= \lambda_{1}\mathcal{L}_{cm}+\lambda_{2}\mathcal{L}_{rm},
\end{equation}
where $\lambda_{1}$ and $\lambda_{2}$ represent the  coefficients of ${L}_{cm}$, and ${L}_{rm}$. 

For the kernel segmentation task, the binary cross-entropy (BCE)  loss and dice loss are used to optimize it.
To balance the ratio of positive and negative samples, we adopt hard negative mining in BCE loss as follows:
\begin{equation}\label{2}
		\mathcal{L}_{BCE}= \sum\limits_{p\in S}-y_p*log(x_p)-(1-y_p)*log(x_p),
\end{equation}
where $y_p$ and $x_p$ represent the ground truth and prediction of the kernel map. The ratio of positive and negative training samples in the chosen set $S$ is 1:3.

Another option dice loss is formulated as follows:
\begin{equation}\label{2}
		\mathcal{L}_{dice}=  \frac{2 \times \sum_{p}(x_p \times y_p)}{\sum_{p}x_p^2 +y_p^2}.
\end{equation}
As we adopt types of loss functions and have two segmentation results needed to optimize, four schemes are performed. 
The corresponding results are shown in Table \ref{tab_loss1}, and the best scheme uses BCE loss twice. Notably, the coarse mask is one-fourth the size of the ground truth, necessitating quadruple upsampling before loss calculation.

\subsection{Inference}
During the inference stage, the refined mask is binarized first, and some tiny positive regions are abandoned. The other regions are generated kernel contours through some morphological operation. The kernel contour is expanded with an offset $O$ to generate text bounding boxes. The offset of $i$th kernel can be calculated as:
\begin{equation}\label{2}
	{O}_{i} = \frac{\hat{A}_i\times \beta}{\hat{P}_i},
\end{equation}
where $\hat{P}_i$ and $\hat{A}_i$ are the perimeter and area of $i$th kernel. $\beta$ is expansion factor, which is set 1.5.
%
%
%
%
%
%
%
\begin{table*}[!t]
	\center
	
	 {
		\caption{  Ablation study on the effect of SCM and MIEM on detection performance on the ICDAR2015, MSRA-TD500, Total-Text, and CTW1500.  }
		\centering
		{ 
			\begin{tabular}{ccc|ccc|ccc|ccc|ccc}
				
				\hline
				\multirow{2}{*}{Kernel}  
				&\multirow{2}{*}{SCM} &\multirow{2}{*}{MIEM }
				& \multicolumn{3}{c}{MSRA-TD500}	        &\multicolumn{3}{c}{ICDAR2015}  	 &\multicolumn{3}{c}{Total-Text}	        &\multicolumn{3}{c}{CTW1500} \\    
				\cline{4-15} 
				&	& &P &R &F   &P &R &F  &P &R &F  &P &R &F  \\ \hline

				\checkmark &$\times$ &$\times$ &82.3 &76.6 &79.4   &87.9 &77.2 &82.2 &84.5 &78.1 &81.2   &86.7 &79.9 &83.1  \\ 
				\checkmark &\checkmark &$\times$  &81.4 &81.7 &81.6   &86.0 &81.6 &83.7 &85.4 &80.6 &83.0   &86.6 &82.0 &84.2 \\ 
				
				\checkmark   &\checkmark &\checkmark &89.2 &80.6&\textbf{84.7}  &88.7 &80.5 &\textbf{84.4}	&87.1 &82.5&\textbf{84.7}  &87.3 &82.3 &\textbf{84.7} \\ \hline
			\end{tabular}
			
		}
		\label{tab_abs1}	
		
	}
\end{table*}

%
%
%
%
\section{Experiment}
\label{experiment}
In this section, we first introduce the used datasets. Then, the adopted evaluation metrics and the  implementation details are presented. Subsequently, we conduct ablation studies on CTW1500, MSRA-TD500, TotalText, and ICDAR2015. In addition, we compare existing state-of-the-art methods with our method on the above datasets to demonstrate the superiority of the STD.   The detection results based on the baseline and the proposed method are compared. The cross-dataset experiments further prove the shape robustness of the proposed method. Finally, we visualize the limitations of STD  and analyze the corresponding reasons.      

\begin{table}
	
{\center
		{	
			\caption{  The detection performance of STD with different feature enhance module  on four public benchmarks.}
			\begin{tabular}{cccccc}
				
				\hline Datasets  &Module &F &Para.(M) & Gflops & FPS\\
				\hline \multirow{3}{*}{Total}  &FPEM + FFM &84.2 &12.2 & 44.3&20.5\\
				&FPEM$\times$2 + FFM  &84.2 &12.3 &45.4 &19.7\\
				&MIEM &\textbf{84.7} &12.4 & 45.5 &20.8\\
				\hline \multirow{3}{*}{CTW} &FPEM + FFM &83.9  &12.2 &28.3 &33.8\\
				&FPEM$\times$2 + FFM  &83.9 &12.3 &29.1 &28.9\\
				&MIEM &\textbf{84.7} &12.4 & 29.1 &31.8 \\ 
				\hline \multirow{3}{*}{TD500}  &FPEM + FFM &81.5 &12.2 &37.5 &26.4\\
				&FPEM$\times$2 + FFM  &83.0 & 12.3&38.5 &23.7\\
				&MIEM &\textbf{84.7} &12.4  &38.5 &26.5\\
				\hline \multirow{3}{*}{IC15}  &FPEM + FFM &84.1 &12.2 &37.5 &19.5\\
				&FPEM$\times$2 + FFM  &\textbf{84.4} &12.3 &38.5 &17.5 \\
				&MIEM &\textbf{84.4} &12.4 & 38.5 & 19.3\\
			
				\hline
		\end{tabular}
		\label{tab_fpemm}
	}}
	
\end{table}
  
\subsection{Datasets}
\textbf{CTW1500}~\cite{yuliang2017detecting} consists of 1,500 samples, 1,000 for training and 500 for testing. Specifically, it contains lots of curved instances that are labeled with line-level.  

\textbf{ICDAR2015}~\cite{karatzas2015icdar} includes 1,000 training images and 500 testing images which enjoy a more complicated background. Moreover, the local-level features of the background are similar to the text regions, which brings challenges to detection.  

\textbf{Total-Text}~\cite{ch2017total} contains many curved texts with word-level labels, which bring challenges to the model. There are 1,255 images for training and 300 images for testing.      

\textbf{SynthText}~\cite{gupta2016synthetic} includes 800k training images which are 
synthesized by various texts and scene images.   It is mainly used to pre-train to improve the robustness of the module.    

\textbf{ICDAR2017 MLT} \cite{icdar2017} is a  large-scale text dataset that contains nine language texts that bring tough challenges. It contains 7,200 training images, 1,800 validation images, and 9,000 testing images.

\textbf{MSRA-TD500}~\cite{yao2012detecting} is composed of 500 images, 300 for training and 200 for testing. It mainly contains line-level instances, and we follow previous methods \cite{db, db++, pan, cm, rsmtd, pan++} to introduce 400 images of HUST-TR400 \cite{yao2014unified} as the extra training set.  

\subsection{Evaluation Metrics}     
To achieve a fair comparison, following the previous works \cite{db}, \cite{db++}, \cite{pan}, we adopt precision (P), recall (R), F-measure (F), and FPS to evaluate the performance, where F-measure is calculated by precision and recall. F-measure and FPS are used to evaluate the detection accuracy and speed in the following experiments. TP, FP, and FN represent the number of true positive, false positive, and false negative samples. Specifically, P, R, and F can be represented as follows:

 		\begin{align}
 		P= \frac{TP}{TP + FP} ,
 		\\
 		R= \frac{TP}{TP + FN} ,
 		\\
 		F= \frac{2 \times P \times R}{P + R} .
 		\end{align}


\subsection{Implementation Details}   
The backbone of STD is ResNet \cite{resnet} with deformable \cite{deformable2} convolution. Feature Pyramid Network (FPN) is utilized to fuse the different scale feature maps. The three training strategies are adopted: (1) Pre-training on  SynthText for four epochs. (2) Pre-training on ICDAR2017MLT for 400 epochs. (3) Without pre-training on extra datasets.   Our STD is trained in 1200 epochs with an initial learning rate of 0.007. The learning rate is adjusted by the "poly" strategy \cite{zhao2017pyramid}, and the stochastic gradient descent (SGD)  is adopted.   Moreover, we set the weight decay and momentum as 0.0001 and 0.9, respectively.   During the training stage, the images are resized to 640 $\times$ 640 with random cropping, flipping, and rotation. The coefficients of the loss  $\lambda_{1}$ and $\lambda_{2}$ are 6 and 1, respectively. The parameter $\alpha$ is initialized to -1. All the speeds referred to in this paper are tested on a GTX 1080Ti GPU and an i7-6800K CPU.             
\subsection{Ablation Study}  
\begin{figure}[t]
	\centering
	\includegraphics[width=1.0\linewidth,scale=0.95]{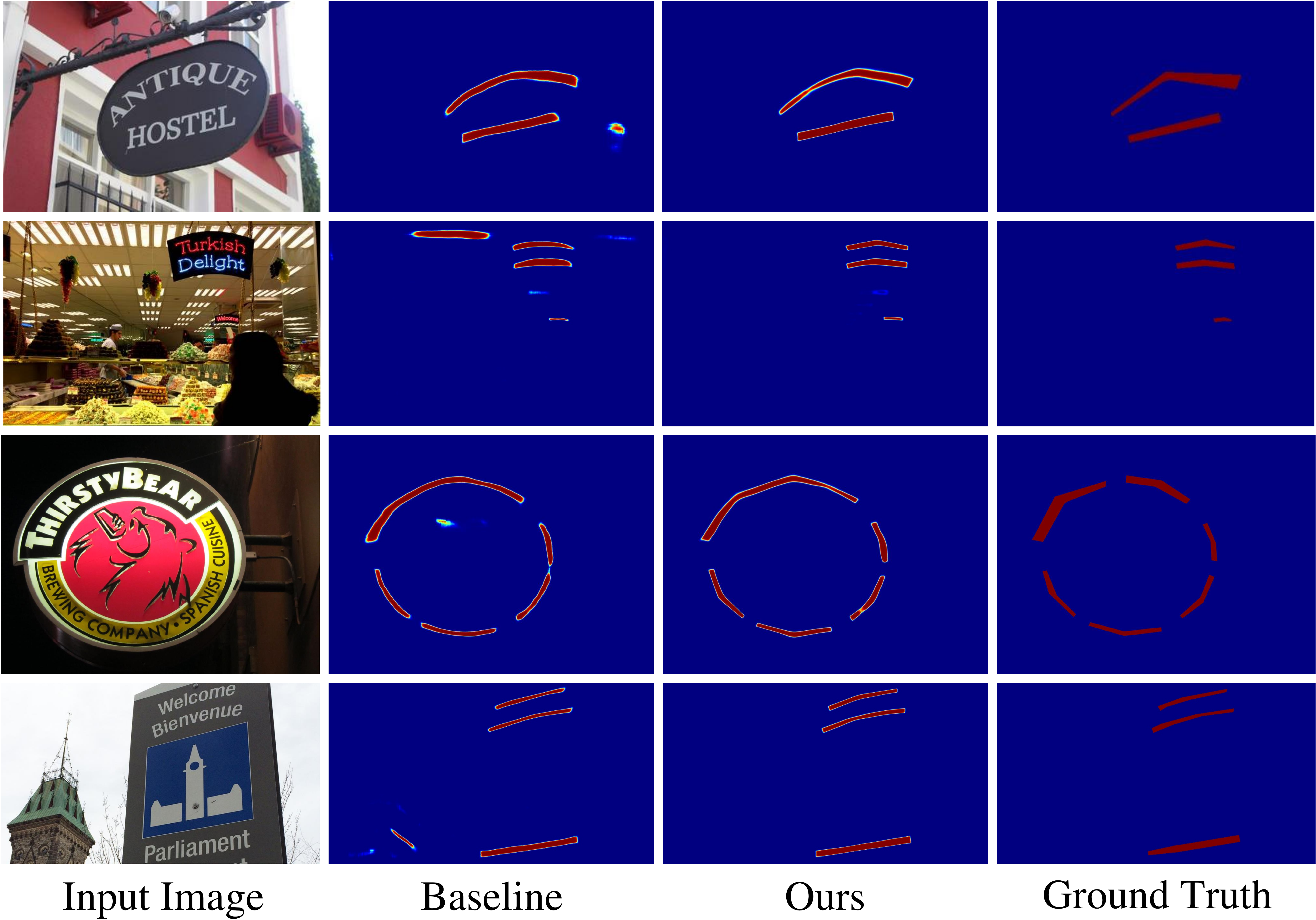}
	\caption{ 
		The visualization of the proposed method and the baseline (only predict text kernels). The ground truth and prediction on the image refer to text kernels. Compared to the former, the latter removes some impurities and generates accurate predictions to achieve more reliable results.}
	\label{vis_second}
\end{figure}

To demonstrate the superiority of the proposed STD, the ablation studies are conducted on multiple public datasets. For a fair comparison, no additional datasets are introduced for pre-training on ablation experiments. The detection results of baseline and our STD are visualized in Fig. \ref{vis_second}. Compared with the baseline, the proposed STD calibrates the prediction to remove some false positive samples.


\begin{table*}[!t]
	\center
	
 {
		\caption{  Ablation study on the choice of loss on detection performance on the ICDAR2015, MSRA-TD500, Total-Text and CTW1500.   ``BCE'' and ``DICE'' represent the binary cross-entropy loss and dice loss.}
		\centering
		{ 
			\begin{tabular}{cc|ccc|ccc|ccc|ccc}
				
				\hline
			 \multirow{2}{*}{Coarse}  &\multirow{2}{*}{Refined}
				& \multicolumn{3}{c}{MSRA-TD500}	        &\multicolumn{3}{c}{ICDAR2015}  & \multicolumn{3}{c}{TotalText}	        &\multicolumn{3}{c}{CTW1500}\\    
				\cline{3-14} 
				& &P &R &F  &P &R &F &P &R &F  &P &R &F \\ \hline

				BCE &BCE &87.8 &79.4 &\textbf{83.4}  &88.7 &80.5 &\textbf{84.4}  &87.1 &82.5 &\textbf{84.7}   &87.3 &82.3 &\textbf{84.7} \\ 
				BCE &DICE  &82.4 &81.8 &82.1  &87.5 &80.1 &83.7   &86.9 &82.5 &84.6  &86.1 &82.7 &84.4 \\ 
				DICE &DICE  &83.3 &82.5 &82.9  &86.0 &80.7 &83.3   &84.8 &83.1 &84.0  &86.0 &82.2 &84.1\\ 
				DICE &BCE &81.1 &82.5 &81.8  &87.0 &81.1 &84.0  &85.6 &83.0 &84.3  &86.0 &83.2 &84.6	\\ \hline
				
			\end{tabular}
			
		}
		\label{tab_loss1}	
		
	}
\end{table*}

%
%
%
%
%
%

\begin{table}
	
	\center
{
		\caption{ The detection performance of STD with different pre-training conditions on four public benchmarks.}
		\begin{tabular}{ccccc}
			
			\hline Datasets &Ext. & P &R &F \\
			\hline \multirow{3}{*}{TotalText} &None &87.1 &82.5 &84.6 \\
			&SynthText & 89.1 & 83.0 &\textbf{85.9}\\
			&ICDAR2017 & 87.9 & 83.9 &85.8\\
			\hline \multirow{3}{*}{MSRA-TD500} &None &89.2 &80.6 &84.7 \\
			&SynthText & 92.2 &83.2 &87.4\\
			&ICDAR2017 & 91.0 & 85.1 &\textbf{87.9}\\
			\hline \multirow{3}{*}{CTW1500} &None &87.3 &82.3 &84.7 \\
			&SynthText &88.7 & 84.1 &\textbf{86.3}\\
			&ICDAR2017 & 87.9 & 84.3 &86.1\\
			\hline \multirow{3}{*}{ICDAR2015} &None &88.7 &80.5 &84.4 \\
			&SynthText & 88.6 & 80.6 &84.4\\
			&ICDAR2017 & 89.3 & 81.1 &\textbf{85.0}\\
			\hline
		\end{tabular}
		\label{tab_pre}
	}
	
\end{table}

\begin{table}
	
	\center
	{
			
		\caption{  The detection performance and speed of different schemes  on ICDAR2015 and CTW1500.}
	
		\begin{tabular}{llllll}
			
			\hline Datasets &Methods  &F & Gflops & FPS\\
			\hline\multirow{3}{*}{ICDAR2015}  &Baseline &82.2  & 32.1 &26.0\\
			&Baseline + SCM  &83.7 &35.3 &23.5\\
			&Baseline + SCM + MIEM &\textbf{84.4} & 38.5 &19.3\\ 
			
			\hline\multirow{3}{*}{CTW1500}  &Baseline &83.1  & 24.3 &44.8\\
			&Baseline + SCM  &84.2 &26.7 &39.2\\
			&Baseline + SCM + MIEM &\textbf{84.7}  & 29.1 &31.8\\ 
			
			\hline
		\end{tabular}
		\label{tab_fps}

	}
	
\end{table}
\subsubsection{Effectiveness of the SCM}
\begin{figure}[t]
	\centering
	\includegraphics[width=1.0\linewidth,scale=1.0]{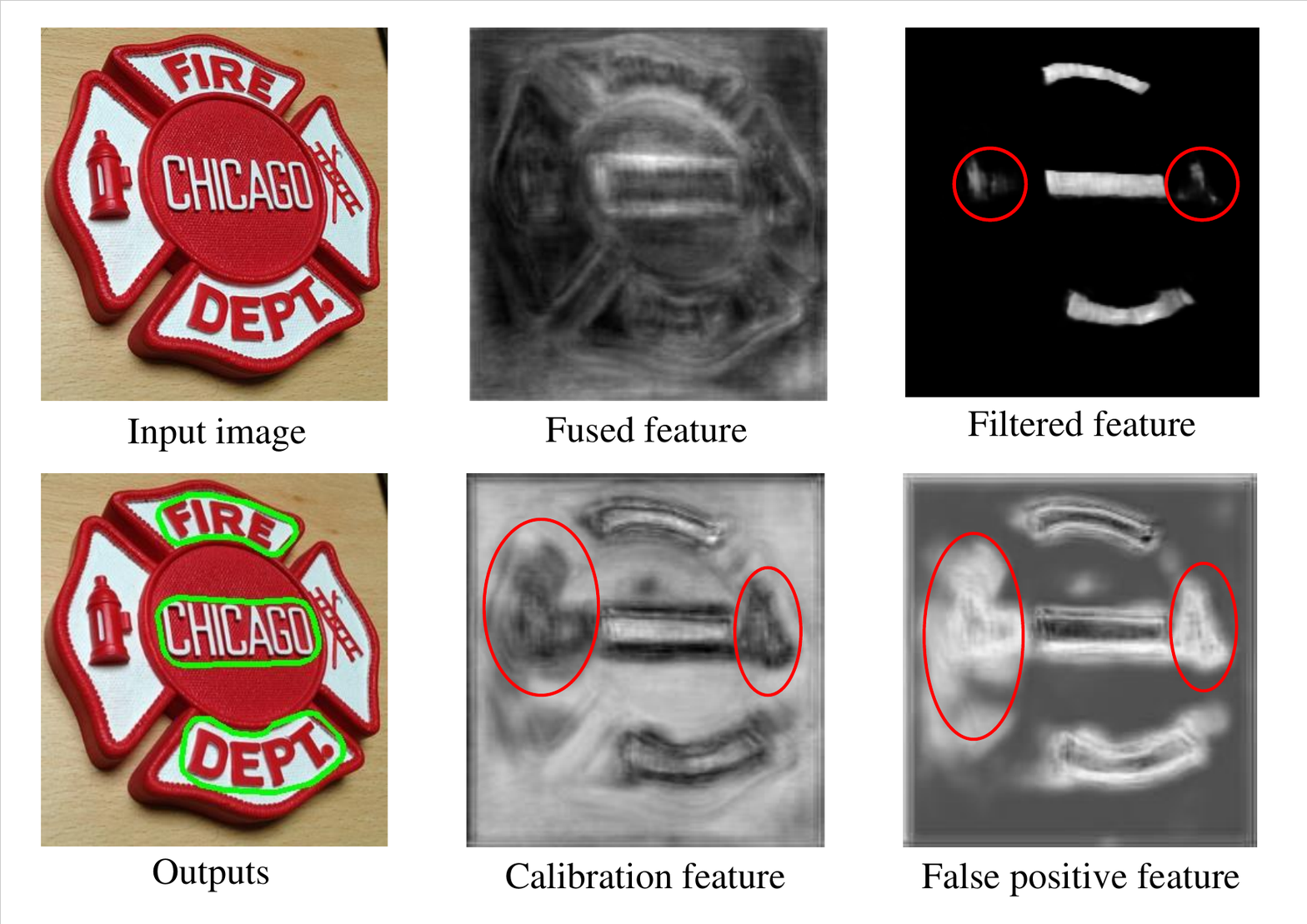}
	\caption{
		The visualization of the different feature map.}
	\label{feature}
\end{figure}

As mentioned above, the spotlight calibration module concentrates efforts on the candidate region based on the coarse mask. Moreover, it calibrates the predictions of the text kernel accurately to eliminate some false positive samples, effectively improving detection performance.  In this section, we conduct ablation experiments to verify the superiority of SCM on CTW1500, MSRA-TD500, ICDAR2015, and Total-Text. As shown in Tab. \ref{tab_abs1}, benefiting from the SCM verified the coarse mask, it is  2.2$\%$ and 1.5$\%$ higher than the baseline in performance on multi-directional datasets MSRA-TD500 and ICDAR2015, respectively. For the irregular-shaped datasets Total-Text and CTW1500,  our method achieves 1.8$\%$ and 1.1$\%$ gains, respectively. The above experiments prove the effectiveness of the SCM. All experiments in Table \ref{tab_abs1} adopt ResNet18 as the backbone. To further prove the validity of SCM, we visualize different feature maps mentioned in this paper. In the model, $\alpha$ is updated to be -0.95. As shown in Fig. \ref{feature}, the filtered feature map predominantly comprises positive features. The false positive feature focuses on some hard negative regions, including text edge areas and areas with textures similar to text. The calibration feature map suppresses these regions prone to misjudgment to obtain the final refined results.

\subsubsection{Influence of the MIEM}
The multivariate information extraction module (MIEM) is proposed to address the variations of scene texts. To be specific, it extracts geometric features from multiple perspectives to cope with the diversity of texts in shape, scale, and orientation. We conduct ablation experiments to demonstrate the superiority of MIEM on ICDAR2015, MSRA-TD500, TotalText, and CTW1500. As shown in Table \ref{tab_abs1},  it brings 3.1 $\%$ and 0.7 $\%$ improvement in performance on MSRA-TD500 and ICDAR2015, respectively.  In addition, our method bring 1.7 $\%$ and 0.5$\%$ gains on curved text datasets Total-Text and CTW1500, respectively. The above experiments prove the effectiveness of the MIEM.
Moreover, we compare the MIEM with FPEM and FFM \cite{pan}, which are proposed in PAN \cite{pan} on the above public datasets. As shown in Table \ref{tab_fpemm}, the proposed STD equipped with MIEM improves 0.5$\%$ and 0.5$\%$ than ``FPEM + FFM'' and ``FPEM $\times$ 2 + FFM'' on Total-Text. For CTW1500 and MSRA-TD500, compared to ``FPEM + FFM'', our method brings 0.8$\%$ and 3.2$\%$ gains, respectively. In addition, compared to ``FPEM $\times$ 2 + FFM'', our method improves F-measure by 0.8$\%$ and 1.7 $\%$ on CTW1500 and MSRA-TD500, respectively. For  ICDAR2015, compared to  ``FPEM + FFM'', the proposed MIEM improves F-measure by 0.3$\%$, which is equal to ``FPEM $\times$ 2 + FFM''. The above experiments also demonstrate the superiority of the proposed MIEM. In terms of computational and parameter complexity, the proposed MIEM is comparable to ``FPEM $\times$ 2 + FFM''. Benefiting from the parallel design of MIEM, compared to ``FPEM $\times$ 2 + FFM'', it achieves a slight speed advantage while enhancing performance. 

\begin{figure*}[t]
	\centering
	\includegraphics[width=0.9\linewidth,scale=1.0]{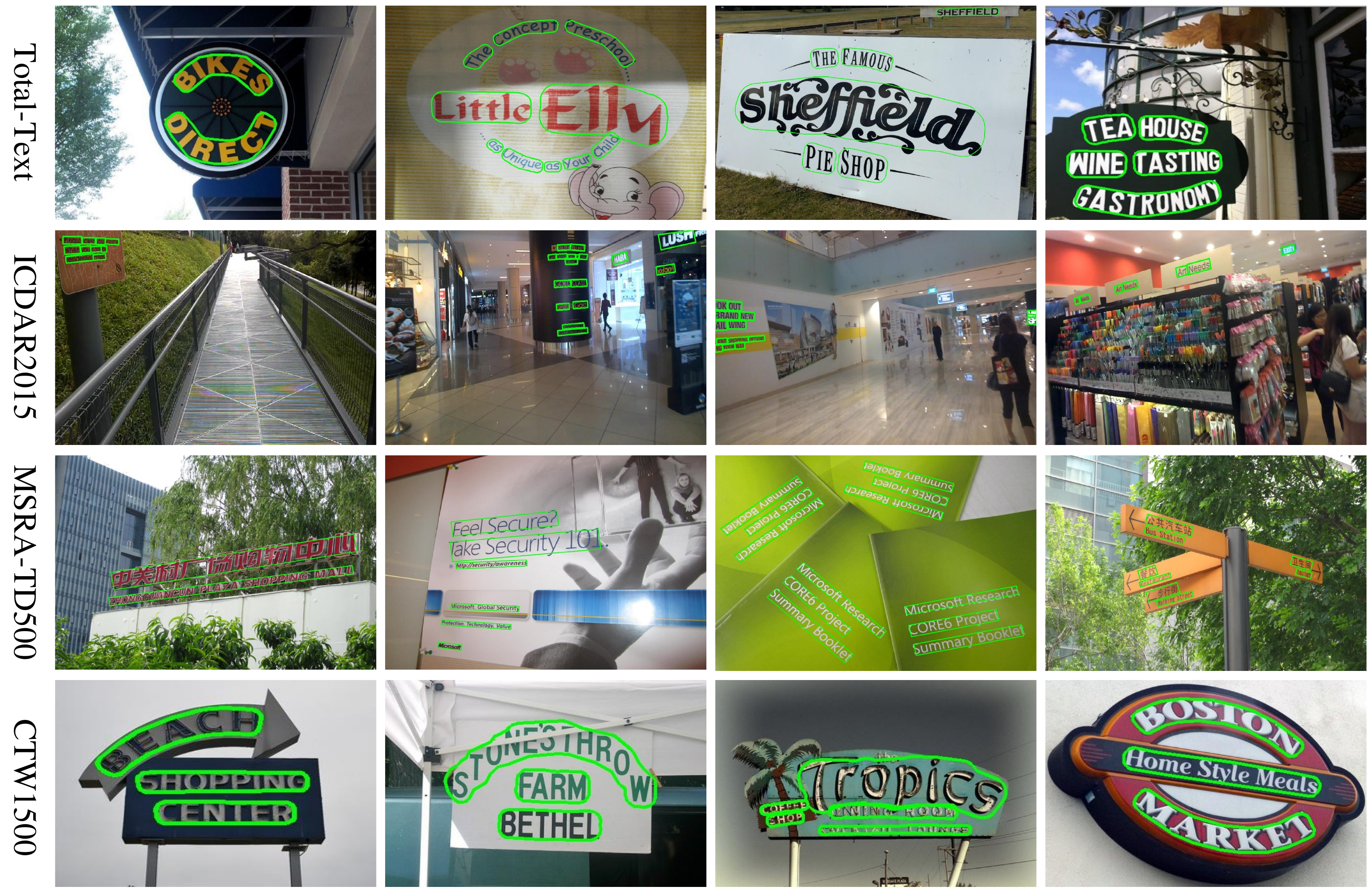}
	\caption{
		Some visual results of the proposed STD. From top to bottom, the samples are selected from Total-Text, ICDAR2015, MSRA-TD500, and CTW1500. }\label{vis}
\end{figure*}

\subsubsection{Influence of the loss}
To verify the influence of different losses, we use BCE loss and dice loss to design four schemes that are conducted on multiple public benchmarks. As shown in Table \ref{tab_loss1},  when the coarse mask and refined mask both adopt BCE loss, the best detection performances are 83.4$\%$, 84.4$\%$, 84.7$\%$, and 84.7$\%$ in terms of F-measure, which are achieved on MSRA-TD500, ICDAR2015, CTW1500, and Total-Text, respectively. When the coarse mask and refined mask both adopt dice loss, compared with the first scheme, the detection performance on ICDAR2015, CTW1500, and Total-Text is significantly limited. For the MSRA-TD500, its F-measure is suboptimal among the four schemes. When the coarse mask adopts BCE loss, and the refined mask adopts dice loss, compared with the best results, the F-measure decreases 1.3$\%$, 0.7$\%$, 0.1$\%$, and 0.3$\%$ on MSRA-TD500, ICDAR2015, Total-Text, and CTW1500, respectively. When the coarse mask adopts dice loss and the refined mask adopts BCE loss, the F-measure has dropped 1.6$\%$, 0.4$\%$, and 0.4$\%$ on MSRA-TD500, ICDAR2015, and Total-Text, respectively. For the CTW1500 dataset, the detection performance is almost the same as the first scheme. Based on the above experimental results, we can conclude that arbitrary-shaped datasets are less affected by loss function, whereas multi-directional datasets are highly influenced by the choice of loss. Dice loss tends to focus on the intersection of regions, while BCE loss emphasizes pixel-level deviations. Multi-directional text datasets, with text contours post-processed as quadrilaterals, are easily influenced by the prediction of the network. In contrast, text in arbitrarily shaped datasets is post-processed as polygons, which are more resistant to interference.

\begin{table*}[ht]
	\center
	\renewcommand\arraystretch{1.0}
	\setlength{\tabcolsep}{2.5mm}
	
	\caption{  Comparison with existing state-of-the-art (SOTA) approaches on the MSRA-TD500 and CTW1500 datasets. ``\textcolor{red}{\textbf{Red}}'', \textcolor{blue}{\textbf{Blue}}'' and  ``\textcolor{green}{\textbf{Green}}'' represent the optimal, sub-optimal and  the third best performance, respectively. ``CTW/Art'' represents that the results of MSRA-TD500 and CTW1500 are pre-train on CTW1500 and ICDAR2019Art. ``MLT'' and ``Synth'' represent the method pre-training on ICDAR2017MLT and SynthText. ``Synth+'' represent the method pre-training on SynthText 150K. ``Synth++'' represent the method pre-training on the mixture of SynthText 150K, MLT
		and Total-Text.  ``*'' and ``$\star$'' represnet the speed is tested on the RTX 3090 and RTX TITAN, respectively.}
	
	{ 
		\begin{tabular}{cccc|cccc|cccc}
			
			\hline
			
			\multirow{2}{*}{Methods} &\multirow{2}{*}{Venue} &\multirow{2}{*}{Ext.} &\multirow{2}{*}{Backbone }
			& \multicolumn{4}{c}{MSRA-TD500}	        &\multicolumn{4}{c}{CTW1500}  \\    
			\cline{5-12} 
			& & & &P &R &F &FPS &P &R &F &FPS\\ \hline 
			PSE-1s\cite{pse}  &CVPR’19&MLT &ResNet50&- &- &- &- &84.8 &79.7 &82.2 &3.9\\ 
			PAN\cite{pan} &ICCV’19 &Synth &ResNet18 &84.4 &83.8 &84.1 &30.2 &86.4 &81.2 &83.7 &39.8\\ 
			DRRG\cite{drrg} &CVPR’20 &MLT &VGG16 &88.1 &82.3 &85.1 &- & 85.9 &83.0 &84.5 &- \\

			ContourNet\cite{contournet}  &CVPR'20 &- &ResNet50 &- &- &- &- &84.1 &83.7 &83.9 &4.5\\ 
			
			DBNet\cite{db} &AAAI’20&Synth&ResNet50 &{91.5} &{79.2} &{84.9} &{32} &{86.9} &{80.2} &{83.4} &{22}\\ 
			DBNet\cite{db} &AAAI’20&Synth&ResNet18 &{90.4} &{76.3} &{82.8} &62 &{84.8} &{77.5} &{81.0} &55\\

			CTNet\cite{ct}&NeurIPS’21 &Synth &ResNet18 &90.0 &82.5  &86.1 &34.8 &{88.3} &79.9 &83.9 &40.8\\ 
			PCR\cite{pcr}  &CVPR’21 &MLT &DLA34 &90.8 &83.5 &87.0 &- &87.2 &82.3 &84.7 &-\\ 
			
			ReLaText \cite{relatext} &PR'21 &Synth &ResNet50 &90.5 &83.2 &86.7 &8.3 &86.2 &83.3 &84.8 &10.6\\
			TextBPN\cite{textbpn}&ICCV’21 &Synth  &ResNet50   &85.4 &80.7 &83.0 &12.7 &87.8 &{81.5} &84.5 &12.2 \\ 
			TextBPN\cite{textbpn}&ICCV’21 &MLT &ResNet50   &86.6 &84.5 &85.6 &12.3 &86.5 &{83.6} &85.0 &12.2 \\  	
			LPAP\cite{lpap} &TOMM'22 &Synth &ResNet50 & 87.9 &77.7 &82.5 &- &84.6 &80.3 &82.4 &- \\
			
			ASTD \cite{astd} &TMM'22 &- &ResNet101 &- &- &- &- &87.2 &81.7&84.4 &- \\
			PAN++ \cite{pan++} &TPAMI'22 &Synth &ResNet18 &85.3 &84.0 &84.7 &32.5 &87.1 &81.1 &84.0 &36.0\\
			RP-Text \cite{rp} &TMM'22 &Synth &ResNet18  &88.4 &84.6 &86.5 &27.3 &87.8 &81.6 &84.7 &23.8 \\
			
			
			DC \cite{dc} &PR'22 &Synth &- &87.9 &83.1 &85.4 &- &86.9 &82.7 &84.7 &-\\
			CMNet\cite{cm} &TIP’22 &- &ResNet18 &89.9 &80.6 &85.0 &41.7 &86.0 &82.2 &84.1 &50.3\\  
			FCBBT \cite{tang2022few} &CVPR’22 &Synth+  &ResNet50  &{91.6} &{84.8} &88.1 &{-} &{88.1} &{82.4} &85.2 &{-}\\ 
			DPText-DETR\cite{dptext} & AAAI'23 &Synth++ &ResNet50 &- &- &- &-& {91.7} & 86.2 & \textcolor{red}{\textbf{88.8}} & - \\
			ZTD \cite{zoom} &TNNLS'23 &Synth &ResNet18 &91.6 &82.4 &86.8 &59.2 &88.4 &80.2 &84.1 &76.9 \\
			FS\cite{fs}&TIP'23&CTW/Art &ResNet18&{90.0} &{80.4} &{84.9} &{35.5} &{84.6} &{77.7} &{81.0} &{35.2}\\ 
			RSMTD \cite{rsmtd} &TMM'23 &Synth &ResNet18  &89.8 &83.1 &86.3 &62.5 &87.8 &80.3 &83.9 &72.1 \\
			FS\cite{fs}&TIP'23&CTW/Art &ResNet50  &{89.3} &{81.6} &{85.3} &{25.4} &{85.3} &{82.5} &{83.9} &{25.1}\\ 
			DBNet++\cite{db++}&TPAMI’23&Synth &ResNet18 &{87.9} &{82.5} &{85.1} &55 &{84.3} &{81.0} &{82.6} &{49}\\
							DBNet++\cite{db++}&TPAMI’23&Synth &ResNet50  &{91.5} &{83.3} &{87.2} &{29} &{87.9} &{82.8} &{85.3} &{26}\\ 
			LRANet \cite{lra} &AAAI'24 &Synth+ &ResNet50 &92.3 &86.3 &\textcolor{green}{\textbf{89.2}} &- &89.4 &85.5 &\textcolor{blue}{\textbf{87.4}} &{37.2}\textsuperscript{*}\\
			KAC \cite{kac} &CVPR'24 &Synth &ResNet50 &93.9 &88.1 &\textcolor{red}{\textbf{90.8}} &18.1\textsuperscript{$\star$} &88.6 &85.4 &\textcolor{green}{\textbf{86.8}} &{19.2}\textsuperscript{$\star$}\\
			\hline
			
			STD &Ours &Synth &ResNet18 &{92.2} &{83.2} &87.4 &33.4
			&{88.7} &{84.1} &86.3 &30.6\\ 	
			STD &Ours &Synth &ResNet50 &{92.8} &{86.9} &\textcolor{blue}{\textbf{89.8}} &13.4
			&{88.5} &{84.9} &86.7 &12.1\\ 
			
			\hline
		\end{tabular}
	}
	\label{td500}
	
\end{table*}

\subsubsection{Influence of the pre-training}
Pre-training is used to improve the robustness of the proposed method. We adopt three training strategies: (1) Pre-training on SynthText. (2) Pre-training on ICDAR2017MLT. (3) Without pre-training. MSRA-TD500  is the dataset most affected by pre-training. As shown in Table \ref{tab_pre}, it brings 2.7$\%$ and 3.2$\%$ improvements with pre-training on SynthText and ICDAR2017MLT, respectively. In contrast, ICDAR2015 is influenced minimally by pre-training. CTW1500 and Total-Text include many arbitrary-shaped text instances, which are improved by 1.6$\%$ and 1.3$\%$ with the pre-training on SynthText. In addition, pretraining  on ICDAR2017MLT bring 1.4 $\%$ and 1.2 $\%$ gains on CTW1500 and Total-Text, respectively. For datasets that contain lots of irregular-shaped texts, using SynthText to pre-train is the same effective as ICDAR2017MLT.

\subsubsection{Analysis of speed and computational complexity}
To evaluate the impact of the proposed modules on computational load and inference speed, we analyze the experimental results on the CTW1500 and ICDAR2015 datasets. As shown in Table \ref{tab_fps}, the SCM and MIEM modules respectively increased computational load by approximately 10$\%$. Regarding inference speed, the SCM and MIEM modules resulted in a decrease of 2.5 FPS and 4.2 FPS, respectively, for the CTW1500 dataset. For the ICDAR2015 dataset, the SCM and MIEM modules led to a decrease of 5.6 FPS and 7.4 FPS, respectively.

\subsection{Comparison with State-of-the-Art Methods}     
To establish the superiority of our proposed method, we conduct comparative analyses with prior studies across four public benchmarks. Specifically, ICDAR2015 and MSRA-TD500 are used to verify the superiority of the proposed model for multi-orientated word-level and line-level text. Total-Text and CTW1500 are employed to validate the model's performance with arbitrary-shaped texts. Detection results from these datasets are illustrated in Fig. \ref{vis}.

\begin{table}[t]
	
	\center
	{
		\caption{  Comparison with existing state-of-the-art (SOTA) approaches on the ICDAR2015. ``\textcolor{red}{\textbf{Red}}'', \textcolor{blue}{\textbf{Blue}}'' and  ``\textcolor{green}{\textbf{Green}}'' represent the optimal, sub-optimal and  the third best performance, respectively.}
		\begin{tabular}{cccccc}

			\hline Method &Backbone  & P & R & F & FPS \\
			\hline 
			
			PixelLink\cite{pixellink} &VGG16  &85.5 &82.0 &83.7 &- \\
			PSE-1s\cite{pse} &ResNet50 & 86.9 & {84.5} & 85.7 & 1.6 \\
			PAN \cite{pan} &ResNet18 & 84.0 &81.9 &82.9 &26.1\\
			TextSnake\cite{textsnake} &VGG16 & 84.9 & 80.4 & 82.6 & 1.1\\
			FS \cite{fs} &ResNet18 &88.1 &77.0 &83.2 &15.3 \\
			FS \cite{fs} &ResNet50 &89.8 &82.7 &86.1 &12.1 \\
			ASTD \cite{astd}  &ResNet101  &88.8&82.6 &85.6 &- \\
			LeafText \cite{leaftext} &ResNet50 &88.9 &82.3 &86.1 &- \\
			Boundary\cite{wang2020all} &ResNet50 & 88.1 & 82.2 & 85.0 & - \\
			FCENet\cite{fcenet} &ResNet50& 90.1 & 82.6 & 86.2 &- \\
			ZTD \cite{zoom} &ResNet18 &87.5 &79.0 &83.0 &48.3 \\
			KPN\cite{kpn} &ResNet50& 88.3 & {88.3} & 86.5 & 6.3 \\
			TextDCT\cite{textdct} &ResNet50& {88.9} & 84.8 & \textcolor{green}{\textbf{86.8}} & 7.5 \\
			
			FCBBT \cite{tang2022few}   &ResNet50 & 91.1 &86.7 &\textcolor{red}{\textbf{88.8}} &- \\
			CM-Net\cite{cm} &ResNet18& 86.7 & 81.3 & 83.9 &34.5 \\ 
			Spotter\cite{8812908} &ResNet50& 85.8 & {81.2} & 83.4 & 4.8 \\
			RP-Text \cite{rp} &ResNet18 &89.0 &82.4 &85.9 &13.7 \\
			LPAP\cite{lpap}  &ResNet50  &88.7 &84.4 &86.5 &-\\ \hline
			STD (Syn)  &ResNet50 & 88.9 & {85.2} & \textcolor{blue}{\textbf{87.0}} & 4.4 \\
			\hline
		\end{tabular}
		\label{ic15}
	}
	
\end{table}

\begin{table}
	\center

	{
		\caption{  Comparison with existing state-of-the-art (SOTA) approaches on the Total-Text. ``\textcolor{red}{\textbf{Red}}'', \textcolor{blue}{\textbf{Blue}}'' and  ``\textcolor{green}{\textbf{Green}}'' represent the optimal, sub-optimal and  the third best performance, respectively.}
		\begin{tabular}{cccccc}

			\hline Methods &Backbone & P & R & F & FPS \\
			\hline 
			
			PSENet-1s\cite{pse} &ResNet50& 84.0 & 78.0 & 80.9 & 3.9 \\
			Boundary\cite{wang2020all} &ResNet50& 85.2 & 82.2 & 84.3 & - \\
			RSMTD\cite{rsmtd} &ResNet18 &88.5 &83.8 &86.1 &70.9\\
			FS \cite{fs} &ResNet18 &88.5 &77.0 &81.1 &33.5 \\
			FS \cite{fs} &ResNet50 &88.7 &77.9 &84.1 &24.3 \\
			ZTD \cite{zoom} &ResNet18 &90.1 &82.3 &86.0 &75.2 \\
			PSE+STKM\cite{wan2021self} &ResNet50 &86.3 &78.4 &82.2 &- \\
			DB\cite{db} &ResNet50 & {87.1} & 82.5 & {84.7} &32 \\
			TextRay \cite{textray} &ResNet50 &83.5 &77.9 &80.6 &-\\
			KPN\cite{kpn} &ResNet50 & {88.7} & 87.1  &87.1 &15 \\
			TextDCT\cite{textdct} &ResNet50& {87.2} & 82.7 & {84.9} & 15.1 \\
			ASTD \cite{astd}  &ResNet101  &85.4&81.2 &83.2 &- \\
			CRAFT \cite{craft} &VGG16 &87.6 &79.9 &83.6 &-\\
			FCBBT \cite{tang2022few}   &ResNet50 & 90.7 &85.7 &\textcolor{blue}{\textbf{88.1}} &- \\
			DBNet++\cite{db++} &ResNet50 &{88.9} &{83.2} &{86.0} &28\\
			NASK \cite{nask} &ResNet50 &85.6 &83.2 &84.4 &8.4 \\
			DPText-DETR\cite{dptext} &ResNet50& {91.8} & 86.4 & \textcolor{red}{\textbf{89.0}} & - \\
			LPAP\cite{lpap}  &ResNet50  &87.3 &79.8 &83.4 &-\\ \hline	
						
			STD (Syn)  &ResNet50& 90.7 & {83.9} & \textcolor{green}{\textbf{87.2}} &12.1 \\
			
			\hline
		\end{tabular}
		\label{total}
	}
	
\end{table}

\textbf{Evaluation on ICDAR2015.}  
To prove the environmental robustness of the proposed STD,  we conduct experiments on ICDAR2015, which contains complicated backgrounds. The presence of multi-scaled and multi-oriented text instances further escalates the text detection challenge. During the inference stage, the short side of the image is resized to 1152. As shown in Table \ref{ic15}, the proposed STD achieves 87.0$\%$ in F-measure. It surpasses the existing SOTA methods KPN \cite{kpn} and FS \cite{fs} by 0.5$\%$ and 0.9$\%$, which mainly benefits the fine-grained calibration of SCM. Although FCBBT\cite{tang2022few} is superior to our method, this is mainly attributed to its utilization of a more comprehensive dataset for pre-training. The above results demonstrate the superior ability to cope with texts in complicated backgrounds. 

\textbf{Evaluation on MSRA-TD500.}
MSRA-TD500 is a multi-directional text dataset that contains English and Chinese. We conduct experiments on it to verify the robustness of the proposed method. During the test stage, the short side of the image is resized to 736. As presented in Table \ref{td500}, existing state-of-the-art methods DBNet++ \cite{db++}, PCR \cite{pcr}] and TextBPN \cite{textbpn} achieve 87.2$\%$, 87.0$\%$ and 85.6$\%$ in terms of F-measure. As the superiority of adaptive fuse multi-scale features and adaptive threshold, DBNet++  surpasses PCR  and TextBPN  0.2$\%$ and 1.6$\%$, respectively. It's noteworthy that PCR  and TextBPN  utilize ICDAR2017MLT for pre-training, which is more effective than SynthText used by DBNet++ and our STD. Different from the above methods that use ResNet50 as the backbone, the proposed STD achieves 87.4$\%$ in F-measure even if it adopts a lightweight backbone ResNet18. It mainly benefits the robustness of the prediction calibration module. When using ResNet50 as the backbone, the proposed STD surpasses the DBNet++ 2.6$\%$ in F-measure. These results demonstrate the effectiveness of the proposed STD for coping with long, multi-orientation texts.

\textbf{Evaluation on CTW1500 and Total-Text.} 
To demonstrate the shape robustness of the proposed STD, we compare it with existing state-of-the-art (SOTA) methods on CTW1500 and Total-Text, which contain many curved text instances. During the test stage, the short side of the image is resized to 800.As shown in Table \ref{td500}, the proposed STD achieves 86.3$\%$ and 86.7$\%$ in F-measure on the CTW1500 dataset when adopting ResNet-18 and ResNet-50, respectively. The proposed method surpasses existing SOTA methods TextBPN \cite{textbpn} and FCBBT \cite{tang2022few} 1.7$\%$ and 1.5$\%$, respectively. Additionally, FCBBT \cite{tang2022few} benefits from pre-training on the robust Bezier Curve Synthetic Dataset \cite{abcnet}, abundant in curved texts, enhancing its detection performance. Although DPText-DETR \cite{dptext} and LRANet \cite{lra} are superior to our method, it is mainly because they use a mixture containing three datasets to pre-train. As KAC \cite{kac} adopts a more robust baseline, it also outperforms our method.  However, despite utilizing the lighter ResNet18, the proposed STD outperforms most SOTA methods that use the heavier ResNet50 backbone, such as DBNet \cite{db} and FS \cite{fs}.


Unlike CTW1500, which is a line-level dataset, Total-Text is a word-level dataset. Text detection methods often mistakenly identify two separate instances as a single entity. As we can see from Table VII, although our method is lower than DPText-DETR \cite{dptext} and FCBBT \cite{tang2022few}, it is still superior to most methods. The primary reason for DPText-DETR's superior performance is its pre-training on a more comprehensive dataset, which combines SynthText 150K \cite{abcnet}, ICDAR2019MLT \cite{icdar2019}, and Total-Text, which enhances detection capabilities. Benefiting the SCM can extract multi-class geometry features, our method surpasses existing SOTA methods KPN \cite{kpn}, DBNet++ \cite{db++}, and TextDCT \cite{textdct} by 0.1$\%$, 1.2$\%$, and 2.3$\%$ in F-measure. The aforementioned results and analyses demonstrate the effectiveness of our proposed method in detecting irregular-shaped texts.

\begin{table}
	
	\center
	{
		\caption{  Two groups (word-level and line-level) cross-dataset evaluations, wheree IC15, Total, TD500, and CTW represent ICDAR2015, Total-Text, MSRA-TD500 and CTW1500 datasets, respectively }
		\begin{tabular}{c|c|c|ccc}
			\hline
			Training &Testing &Methods &P &R & F \\ \hline
			\multirow{3}{*}{IC15}  &\multirow{3}{*}{Total} 
			&Textfield\cite{textfield} &61.5 &65.2 &63.3\\
			&&CM-Net \cite{cm}  & 75.8 & {64.5} &69.7 \\ 
			&&STD(ours) & 80.7 &64.8 & \textbf{71.9} \\ \hline 
			\multirow{3}{*}{Total} & \multirow{3}{*}{IC15} 
			&Textfield\cite{textfield} &77.1 &66.0 &71.1\\
			&&CM-Net \cite{cm}  & 76.5 & {68.1} &72.1 \\ 
			&&STD(ours) & 78.6 &70.0 & \textbf{73.9} \\ \hline
			\multirow{3}{*}{TD500} & \multirow{3}{*}{CTW} 
			&Textfield\cite{textfield} &75.3 &70.0 &72.6\\
			&&CM-Net \cite{cm}  & 77.2 & {69.7} &72.8 \\ 
			&&STD(ours) & 84.2 &70.1 & \textbf{76.5} \\ \hline
			\multirow{3}{*}{CTW} & \multirow{3}{*}{TD500} 
			&Textfield\cite{textfield} &85.3 &75.8 &80.3\\
			&&CM-Net \cite{cm}  & 85.8 & {77.1} &81.2 \\ 
			&&STD(ours) & 84.2 &83.5 & \textbf{83.9} \\ \hline
			
		\end{tabular}
		\label{tab_jiaocha}
	}
\end{table}

\begin{figure*}[t]
	\centering
	\includegraphics[width=1.0\linewidth,scale=0.98]{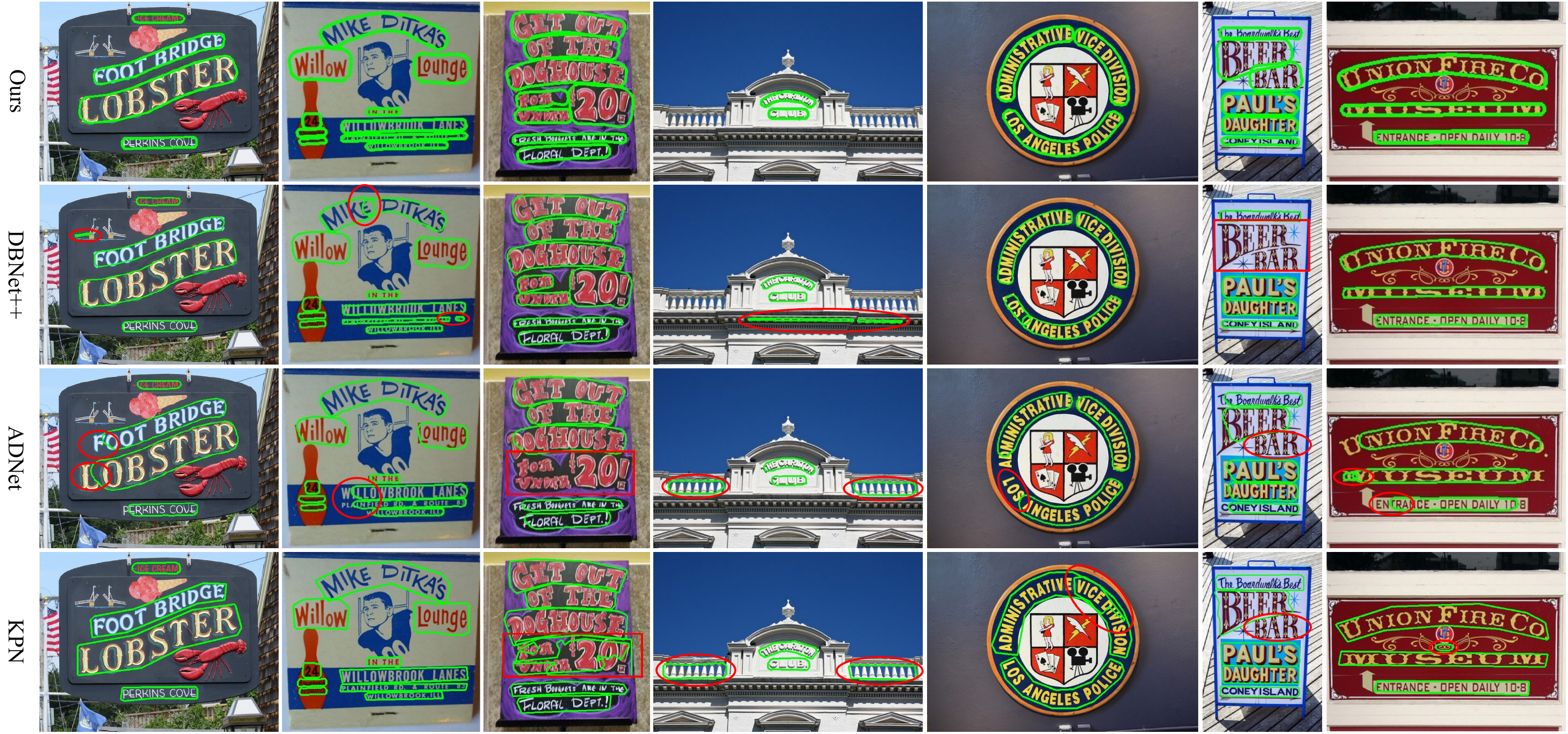}
	\caption{
		The comparison with other state-of-the-art methods. False detections are marked in red.}
	\label{compare_img}
\end{figure*}

\subsection{Cross Dataset Text Detection}  
As shown in Table \ref{tab_jiaocha}, we conduct a multi-condition experiment to show the robustness of the STD. Specifically, the ICDAR2015 and Total-Text are word-level datasets. The MSRA-TD500 and CTW-1500 are line-level datasets. We train on ICDAR2015 and MSRA-TD500 and test on Total-Text and CTW1500. Then, we exchange the training set and the testing set. The proposed method achieves 71.9$\%$ in F-measure when training on  ICDAR2015 and testing on Total-Text. Compared with the  SOTA method CM-Net, the proposed STD surpasses  2.2$\%$ of F-measure. When training on Total-text and testing on ICDAR2015, our STD also achieves 73.9$\%$ of F-measure, a competitive performance. The above experiments powerfully demonstrate the generalization ability of STD. For line-level datasets, we also conduct the cross-train-test experiments. When the proposed method is trained on CTW1500 and tested on MSRA-TD500, it achieves 83.9$\%$ in F-measure, which is superior to some methods (\cite{lpap}, \cite{textbpn}, \cite{db}) directly training on MSRA-TD500. It shows the robustness of STD for long text instances and the adaptability to different scenes.

\subsection{Visual comparsion } 

As shown in Fig. \ref{compare_img}, ADNet \cite{ad} and DBNet++ \cite{db++} misjudge an instance as two, which significantly obstacles the improvement of detection performance. Benefiting from the dual calibration enhancing the ability to recognize whole text instances, the proposed method distinguishes kernel regions and non-kernel regions effectively to alleviate this problem.  In addition,  KPN \cite{kpn}, ADNet \cite{ad}, and DBNet++ \cite{db++} miss some positive samples. In contrast, STD detects these instances accurately with the MIEM, which extracts multiple geometry features. When dealing with a texture resembling the pattern of the scene text, the above methods sometimes misidentify them. In contrast, the proposed STD cognizes them effectively.

\subsection{Comparisons of Coarse Mask and Refined }  
To show the improvement of the refined mask, we compare the detection performance based on the coarse and refined mask, respectively. As shown in Table \ref{first_second}, for ICDAR2015 and MSRA-TD500, the refined mask brings  6.1$\%$ and 6.7$\%$ improvements to the  coarse mask  in F-measure without introducing an extra dataset to assist training. When adopting SynthText to pre-train, the refined mask outperforms the coarse 5.8$\%$ and 3.8$\%$, respectively. For Total-Text and CTW1500, the refined mask brings 3.2$\%$ and 2.5$\%$ F-measure, respectively. With pre-training on SynthText, the improvements become 3.4$\%$ and 2.6$\%$, respectively. Compared to the coarse mask, the refined mask is improved significantly and is more robust. Moreover, compared with the arbitrary-shaped scene text, the multi-oriented scene text is influenced more.

\begin{figure}[t]
	\centering
	\includegraphics[width=1.0\linewidth,scale=0.95]{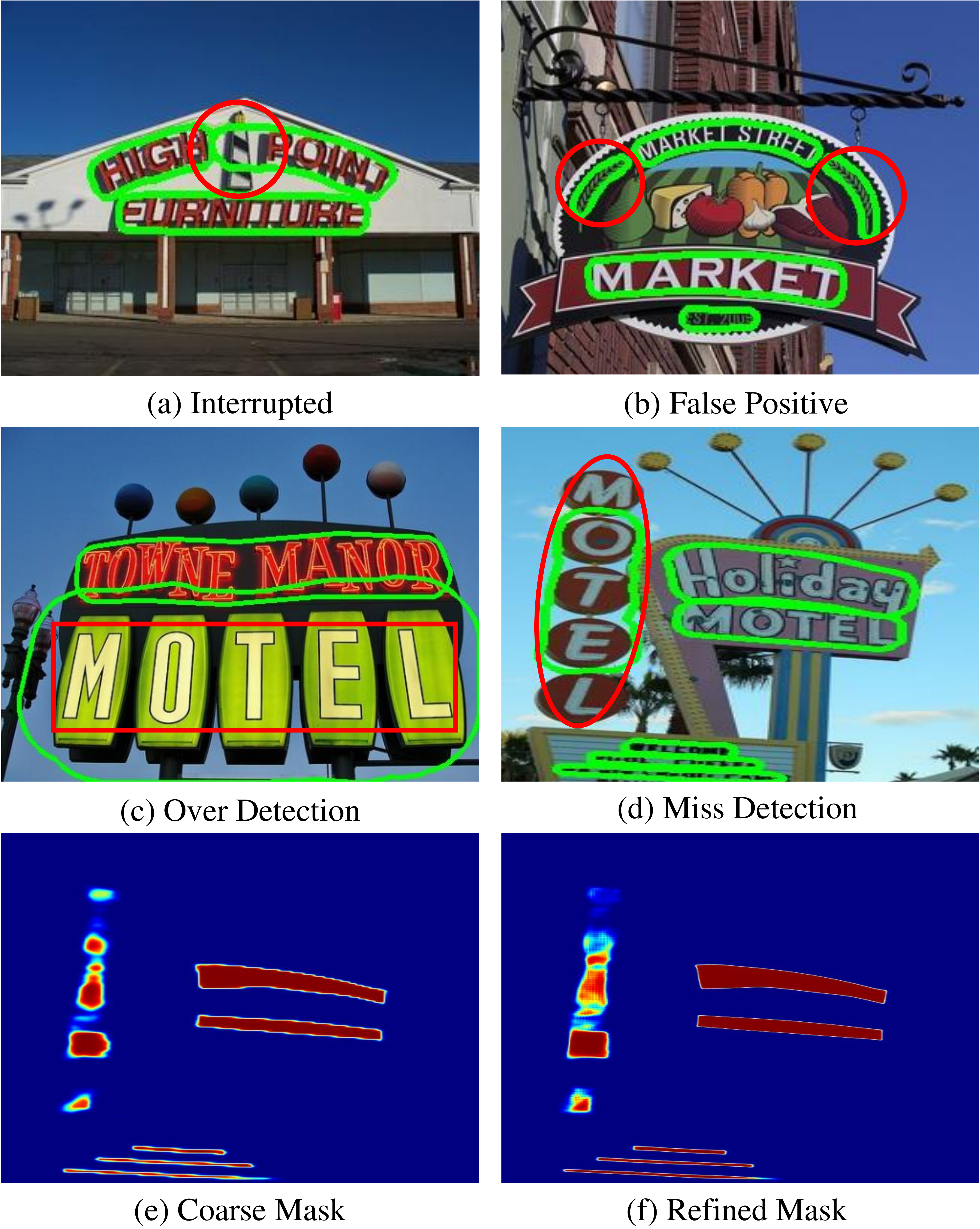}
	\caption{
	Some visualization of error detection. All the misguided predictions are labeled in red. (e) and (f) are the predicted coarse mask and refined mask of (d).}
	\label{limit}
\end{figure}

\subsection{Limitation analysis} 
Fig. \ref{limit}(a) illustrates a text instance mistakenly identified as two separate entities due to obstacle truncation, highlighting the need for enhanced focus on high-level semantic features. Low-level texture features aid in discerning if a pixel is part of text but fail to ascertain whether an obstacle belongs to a text instance. High-level semantic features represent instance-level features, which help the model to determine whether two candidate regions are an entity and whether the backgrounds between them are positive samples. In Fig. \ref{limit}(b), although our method eliminates most false positive samples, due to the vision texture feature not effectively modeling the text feature,  certain text-like patterns are still incorrectly identified, necessitating the use of high-level semantic features to assist the judge. In addition, there is some background that is misjudged as part of the instance in Fig. \ref{limit}(c). The introduction of linguistic features may alleviate this problem well. Fig. \ref{limit}(d) demonstrates incomplete detection of characters from the same instance, attributed to their separation in visual features. As shown in Fig. \ref{limit}(e) and Fig. \ref{limit}(f), the top and bottom part of the instance is miss detected. These characters appear more like independent instances with discontinuous backgrounds compared to other texts. Furthermore, samples of this nature are scarce. We need to focus on instance-level semantic features to assist the prediction further and explore how to detect text effectively based on a few samples.

\begin{table}[t]
	\center
	\renewcommand\arraystretch{1.0}
	\setlength{\tabcolsep}{1.7mm}
	{
		\caption{  Comparison of the results based on the coarse mask and the refined mask on the MSRA-TD500, ICDAR2015, Total-Text, and CTW1500 datasets. }
		
		{ 
			\begin{tabular}{c|c|ccc|ccc}
				
				\hline
				
				\multirow{2}{*}{datasets}&\multirow{2}{*}{Ext.}
				& \multicolumn{3}{c}{Coarse}	        &\multicolumn{3}{c}{Refined}  \\    
				\cline{3-8} 
				&   &P &R &F  &P &R &F \\ \hline

				\multirow{2}{*}{ICDAR2015}  &- &85.7 &72.1 &78.3  &88.7 &80.5 &\textbf{84.4}
				\\ 
				&Syn &85.3 &72.9 &78.6  &88.6 &80.6 &\textbf{84.4} \\ \hline
				\multirow{2}{*}{MSRA-TD500}  &- &82.5 &73.9 &78.0  &89.2 &80.6&\textbf{84.7} \\ 
				&Syn &90.1 &78.0 &83.6  &92.2 &83.2 &\textbf{87.4} \\ \hline
				\multirow{2}{*}{TotalText}  &- &88.1 &75.8 &81.5  &87.1 &82.5 &\textbf{84.7} \\ 
				&Syn &89.7 &76.5 &82.5  &89.1 &83.0 &\textbf{85.9} \\ \hline
				\multirow{2}{*}{CTW1500}  &- &86.6 &78.3 &82.2  &87.3 &82.3 &\textbf{84.7} \\ 
				&Syn &88.3 &79.2 &83.5  &88.7 &84.3 &\textbf{86.1} \\ \hline

			\end{tabular}
		}
		\label{first_second}
	}
\end{table}

\section{Conclusion}

In this paper, an effective spotlight text detector (STD) is proposed, which consists of a spotlight calibration module (SCM) and a multivariate information extraction module (MIEM). The former refines the coarse mask that calibrates the coarse predictions to eliminate some false positive samples.
Specifically,  it focuses on the features of high activation value regions and ignores other regions. The latter extracts multiple geometric features to cope with the diversity of texts in shape, scale, and orientation. Extensive ablation studies verify the effectiveness of the designed modules and prove our method outperforms existing SOTA methods. However, there are still some limitations that need to be solved. In the future, we will explore these problems.
\label{conclusion}

%
%
%

\ifCLASSOPTIONcaptionsoff
\newpage
\fi
\linespread{1}
\bibliographystyle{IEEETran}
\bibliography{IEEEabrv}

\end{document}